# Airplane Type Identification Based on Mask RCNN and Drone Images

**W.T Alshaibani[1], Mustafa Helvaci[1], Ibraheem Shayea[2], Sawsan A. Saad[3], Azizul Azizan[4] and Fitri Yakub[4]**

[1]Institute of Informatics, Satellite Communication and Remote Sensing Program, Istanbul Technical University, ITU Ayazaga Campus, Institute of informatics Building, Sariyer, 34469 Istanbul, Turkey;
[1]Electronics and Communication Engineering Department, Faculty of Electrical and Electronics Engineering, Istanbul Technical University (ITU), 34469 Istanbul, Turkey
[2]Faculty of Engineering and Built Environment, Universiti Sains Islam Malaysia, 71800 Bandar BaruNilai, Negeri Sembilan, Malaysia.
[3]Department of Computer Engineering, College of Computer Science and Engineering, University of Ha'il, Saudi Arabia.
[4]Advance Informatics Department, Razak Faculty of Technology and Informatics, Universiti Teknologi Malaysia, 54100 Kuala Lumpur, Malaysia
[5]Electronic system engineering department, Malaysia-Japan International Institute of Technology (MJIIT), Universiti Teknologi Malaysia, 54100 Kuala Lumpur, Malaysia

Corresponding author: W.T Al-Shaibani (e-mail: al-shaibani18@itu.edu.tr).

This research has been produced benefiting from the 2232 International Fellowship for Outstanding Researchers Program of TÜBİTAK (Project No: 118C276) conducted at Istanbul Technical University (İTÜ), and it was also supported in part by Universiti Sains Islam Malaysia (USIM), Malaysia.This research is supported using Industry-International Incentive Grant number PY/2020/05624

**ABSTRACT** For dealing with traffic bottlenecks at airports, aircraft object detection is insufficient. Every airport generally has a variety of planes with various physical and technological requirements as well as diverse service requirements. Detecting the presence of new planes will not address all traffic congestion issues. Identifying the type of airplane, on the other hand, will entirely fix the problem because it will offer important information about the plane's technical specifications (i.e., the time it needs to be served and its appropriate place in the airport). Several studies have provided various contributions to address airport traffic jams; however, their ultimate goal was to determine the existence of airplane objects. This paper provides a practical approach to identify the type of airplane in airports depending on the results provided by the airplane detection process using mask region convolution neural network. The key feature employed to identify the type of airplane is the surface area calculated based on the results of airplane detection. The surface area is used to assess the estimated cabin length which is considered as an additional key feature for identifying the airplane type. The length of any detected plane may be calculated by measuring the distance between the detected plane's two furthest points. The suggested approach's performance is assessed using average accuracies and a confusion matrix. The findings show that this method is dependable. This method will greatly aid in the management of airport traffic congestion.

**INDEX TERMS** Deep learning; Airplane detection; Drone.UAV, Mask RCNN, CNN, Machine learning. Airplan Type ,

## I. INTRODUCTION

The value of travel has resulted in a considerable increase in the number of flights taken throughout the world [1]. As a result, the number of airports, airplanes, and flight routes has risen. This significant rise has the potential to make airport ground traffic management chaotic and unmanageable. To minimize traffic bottlenecks and delays, the massive volumes of worldwide aviation traffic must be monitored and organized.

Object detection is a technique for determining items automatically, without the need for human interaction. This approach was first utilized for defense and military concerns, but it has since been used to a variety of current technological study fields. Object detection is classified as a remote sensing science research project because it may be used to get meaningful information about items without requiring physical touch. Object detection methods have developed, resulting in the introduction of new methodologies (such as the sliding window scanning method). Low-level vision components are used to complete the recognition process. These elements may be detected using detectors that measure the similarity of forms, textures, and edges, among other things. Prior to the introduction of machine learning-based approaches, this had a beneficial impact on the area of object detection methods. Machine learning offers a significant advancement in the field of object identification since it includes approaches like statistical appearances models, wavelet features, and gradient-based representation.

Object detection history may be divided into two categories: classic detection and current detection. All object identification techniques such as Viola-Jones Detectors [2], Histogram of Oriented Gradients (HOG) [3], and Deformable Part-based Model (DPM) [4] rely on handmade characteristics. The Viola algorithm has exceptional skills for detecting human faces in real time. It was thought to be incredibly quick at the time when compared to other algorithms. The scale-invariant feature transform (SIFT) [5] is improved by HOG, whereas DPM is an extension of the HOG detector. Due to the limits in the extraction process of handcrafted features, deep learning is widely used in modern detection approaches. The use of convolution neural networks has been highlighted in certain recent object identification algorithms (CNN). CNN is capable of learning and acquiring a reliable feature extraction. There are several phases to deep learning models. Some, like the Region Convolution Neural Network (RCNN) and the Feature Pyramid Network (FPN), have two stages, while others, like the Single Shot Multi-Box Detector (SSD) or You Only Look Once (YOLO), only have one (YOLO).

Ross Girshick developed the RCNN to avoid the need to pick a large number of areas [6]. For each image, the selected research approach is utilized to restrict the retrieved areas to 2000. These areas are referred to as region proposals. Rather than categorizing a large number of areas, only 2000 are created and submitted to the classifier after being shrunk to fit the neural network. Fast RCNN uses a similar technique to RCNN, but it feeds the input picture immediately into CNN, which generates a convolution feature map that is used to determine the region of suggestions. Fast RCNN [7] makes a significant contribution to the region of the interest pooling layer. Among deep learning models, faster CNN is considered the first real-time model [8]. Faster CNN launched the Region Proposal Network (RPN), and YOLO is the first in-stage based detector in deep learning [9]

Deep learning is changing the way we think about technology, especially now that we have access to huge quantities of computer power. Computational power enables machines to recognize and identify things, resulting in a slew of new applications that have a direct impact on people's lives. Deep learning may be found in a variety of places, including tools, libraries, institutions, government agencies, and research equipment. Complicated problems are addressed as distant sensing tasks with the use of deep learning. Remote sensing applications have recently received significant enhancements, making them more accurate in terms of object detection, particularly deep learning-based object recognition.

Complicated problems are addressed as distant sensing tasks with the use of deep learning [10]. Remote sensing applications have recently received significant enhancements, making them more accurate in terms of object detection, particularly deep learning-based object recognition . Identifying the airplane type gives useful information about the technical and physical needs of each plane at the airport, enabling for a well-organized, highly efficient, and quick service procedure. Based on the findings of the airplane detection procedure, this study presents a reliable method for determining the kind of detected airplane. The length-based approach and the cabin length-based approach are two methods for distinguishing between different types of airplanes.

The surface area-based technique uses the ground resolution and the total number of covered pixels in the detected region to calculate the surface area of every identified airplane. The length-based technique, on the other hand, depends on calculating the distance between any two furthest locations in the detected airplane region to determine the length of any detected plane.

The rest of this paper is organised as follows. Section 2 highlights the literature review of related works. Section 3 explains the methodology and the practical steps involved. Section 4 discusses the results of the evaluation process. Finally, Section 5 concludes the entire paper.

**II BACKGROUND AND RELATED WORK**

Object detection refers to a group of techniques for recognizing and finding things in pictures and movies. The underlying idea behind all of these approaches is that each object class has distinct specific traits that help in categorization, such as detecting loops and sticks consisting of numerous curves and sub-sticks, as illustrated in Figure 1. Complex classes such as faces, buildings, cars, and airplanes have lately been identified using object detection. Object detection is primarily comprised of two steps: categorization and localization. The classification process' goal is to forecast the class of observed items, whereas the localisation process' goal is to pinpoint their position. Object detection is

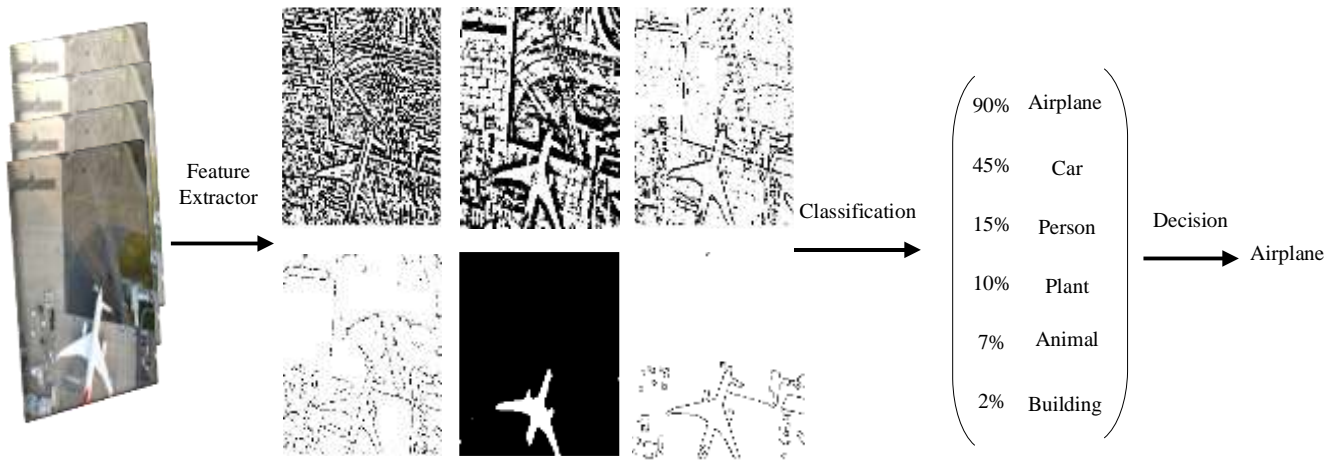

**Figure1:** Airplane detection process.

particularly important in the remote sensing area because of its usefulness in military applications, incursion applications, and border security systems. Machine learning-based techniques and deep learning-based approaches are used to recognize objects [11, 12].

### A. DEEP LEARNING VS MACHINE LEARNING

Because deep learning is largely reliant on data, increasing the amount of data available would improve performance. The performance would be unacceptable if there was insufficient data. Machine learning, on the other hand, is based on constructed rules. Graphics processing units (GPUs) are frequently required for deep learning to work. This is due to the huge quantity of multiplication processes used by the deep learning system. As a result, we may conclude that deep learning is more hardware-dependent than machine learning. Humans must identify and code features in machine learning in terms of domain and data type. Deep learning, on the other hand, streamlines the entire feature extraction procedure for each issue. Machine learning breaks any issue into separate pieces, solves each one individually, and then combines the pieces to obtain the final answer. Deep learning, on the other hand, offers a full answer. However, because there are so many variables, learning takes longer. Deep learning might take days or weeks to complete, whereas machine learning can be completed in seconds or hours at most. The differences between ML and DL are seen in Figure 2. Object detection from a picture is accomplished using a variety of methods. These methods may be divided into two categories: machine learning-based and deep learning-based approaches. Before using a support vector machine to categorize the data, machine learning-based methods must first identify the features. Deep learning, on the other hand, does this without depending on handmade regulations by using a convolution neural network from start to finish. As a result, this study concentrates on deep learning approaches.

### B. DEEP LEARNING METHODS

There are two types of deep learning approaches: supervised and unsupervised. Convolutional neural networks and multilayer perception are two approaches that are supervised. In general, the convolutional neural network collects multi-features from an input picture and sends them to the classifier for labeling. Multilayer perception, on the other hand, transforms the input into vectors for one-class object recognition. Convolutional neural networks contain extensive training procedures that allow them to learn information and make more accurate judgments in the future. Multilayer perception provides a predicted label as a target for the input. Many studies employed supervised techniques, such as Xiang et al. who introduced a hybrid Neural network (HDNN) [11]. Variable receptive blocks caused by HDNN obtain variable-scale features to perform object detection in remote sensing. HDNN is basically a neural network that has a feature map in the last layer divided into blocks with corresponding filter sizes. After identifying the multistage feature, the MLP network can then be used to classify the features. It has been indicated that CNN can correctly recognise locations and HDNN has the ability to learn and distinguish features to recognise objects. In the pixel pair augmentation method, the convolution has been made spatial-spectral for 3-dimensional CNN and 2-dimensional CNN [12, 13]

The unsupervised techniques have the ability to learn feature representations with a number of labels. They do not depend on huge, labelled data as with supervised techniques. There are several unsupervised techniques such as Auto Encoders (AEs) [14], Restricted Boltzmann Machines (RBMs) [14, 15], k-means clustering [16, 17], sparse coding [18-21] and the Gaussian mixture model [22-24]. The unsupervised convolutional neural network has been employed to estimate the weights of the network [21, 25, 26]. Unsupervised spectral-spatial feature learning was represented in the fully residual convolutional neural network. The unsupervised method is suitable for remote sensing applications as the feature can be learned without prior knowledge since remote sensing has limited labelled images. A sparse coding method is used to obtain the dense, low-level features and encode in terms of basic functions to provide a new representation. Figure 3 illustrates the deep learning techniques and related branches.

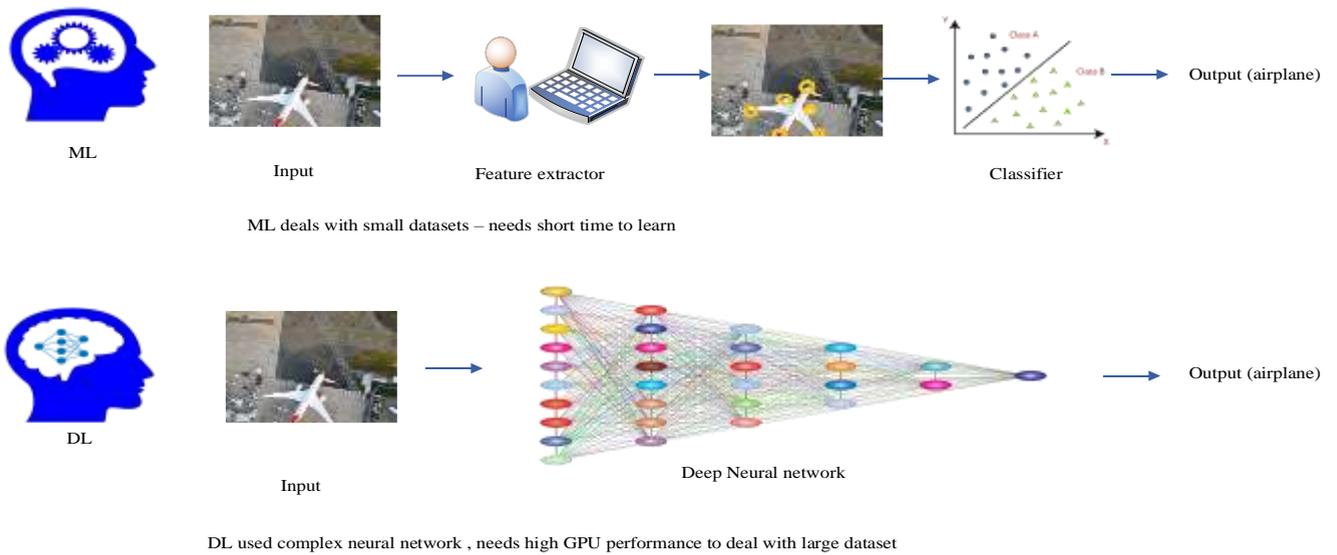

**Figure 2:** **Concept of machine learning ML and deep learning DL**

*C. DEEP LEARNING MODELS*

Region-based Convolutional Neural Networks (R-CNN) [5] utilise an object proposal algorithm called selective search [27] to reduce the number of bounding boxes that are fed to the classifier; approximately 2000 region proposals. The selective search uses local cues (such as texture, intensity and colour) to generate all possible locations of an object. These boxes can be introduced to the CNN based classifier. The fully connected part of CNN takes a fixed-sized input; therefore, all generated boxes should have a fixed size (224×224 for VGG) and feed as the CNN part. The RCNN allows each region in the input image to pass through a ConvNet layer without sharing computations. This action may take significant time, adding additional delays to the SVM classification process. The Fast RCNN [6] forwards the entire input image through the ConvNet and conducts feature extractions at one time. Since the following layers of classification and bounding boxes regressions are completely connected layers that only accept fixed-sized inputs, the extracted features move through a region of interest layer, the (RoI) pooling layer, to obtain fixed sizes. As a result, all features from the entire input image are extracted simultaneously and forwarded to CNN for classification and localisation. This action reduces the delay time compared to RCNN while saving useful storage. Another feature is that Fast RCNN extracts fixed sizes using an RoI pooling layer. Experiments have shown that Fast RCNN improves outcome values; it acquired 66.9% mAP compared to RCNN which obtained 66.0% on PASKAL VOC 2007. The consumed time also decreased to 9.5 hours whereas RCNN consumed 84 hours, signifying a 9-fold reduction in time. In Fast-RCNN, selective search is applied to generate the RoI. The selective search consumes the same running time as the detection network; therefore, it is considered as a slow operation. In Faster-RCNN [8], the selective search has been replaced by a region proposal network (RPN). RPN is a fully convolutional network used to predict proposals with a wide range of scales and aspect ratios. RPN predicts region proposals more efficiently by sharing full-image convolutional networks. It introduces the idea of anchor boxes with different scales (128× 128, 256×256 and 512×512) and different aspect ratios (1:1, 2:1 and 1:2) to handle the variations in the aspect ratio and scale of objects. Faster R-CNN is expanded by the Mask RCNN [30, 31], which replaces RoIPool with RoIAlign and adds a branch for predicting segmentation masks on each Region of Interest (RoI). RoIAlign does not make any adjustments to the input proposal so that it properly fits the feature map. The RoIAlign performs a discrediting process to the floating-number RoI of the feature map. The output of the discrediting process is then divided into quantised spatial bins. Each feature is then covered by a bin using max pooling. The interpolation process is applied to determine the best value for the input feature at four sampled positions in each bin. It simply takes the object proposal and divides it into four equal bins using the bilinear interpolation: top left, top right, bottom left and bottom right. It then applies the average or maximum function to obtain the value. Figure 4 illustrates the deep learning models with further details regarding their functional work, beginning with RCNN, Fast RCNN, Faster RCNN and Mask RCNN

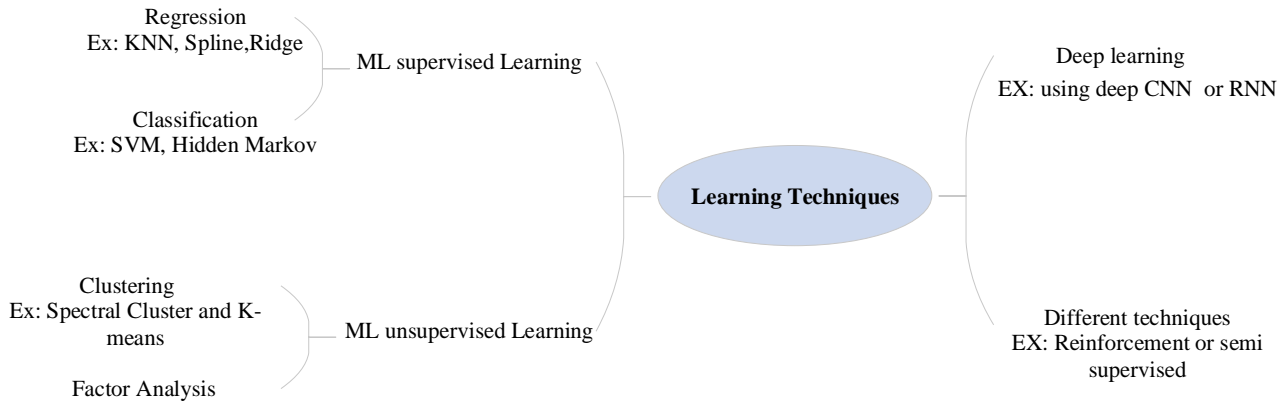

**Figure3:** learning techniques.

## D. RELATED WORK

In [9, 28], the authors employed an enhanced method to detect objects of multi-scales, such as boats in rivers or airplanes. The training data had been collected as elements from large images. The augmentation process was then applied using rotations and random scaling of saturation. The authors of [34] utilised high-resolution images of several labelled cars from multiple areas to detect cars using a deep learning algorithm. In [29], the authors introduced a new approach to detect circle items that correspond to oil tank objects in satellite images. Faster RCNN was initially used to extract the region of the oil tank object, then an improved co-segmentation with saliency co-fusion was applied to identify bright oil tanks. In [30, 31], the authors introduced edge-boxes to obtain edge information and CNN for the feature extractor. Gathering CNN with edge-boxes allows target classification of aircraft objects or non-aircraft objects. A transferred deep model based on AlexNet CNN was introduced by Zhou et al. [32]. In [33], the authors used Single Shot Detector (SSD), faster RCNN, and You Only Look Once (YOLO) with satellite dataset to perform airplane detection. The collected data was divided into three sets: train, assess and test. A training process was then generated for the proposed models. In [34], the authors introduced a fast region-based convolutional neural network (R-CNN) method to detect ships from high-resolution remote sensing imagery. In [35], a pixel-based system was proposed for geometrical special images. The aim of this system is to acquire high accuracy maps by preserving geometrical details in images by considering the spatial-context information. A driving texture feature set was developed for context-based satellite image retrieval in [42]. The benchmark of this study included 37 satellite image chips from various satellite instruments. The efficiency of satellite image similarities was assessed and compared. A method for integrating the high-level information of a shape, previously considered as a coarse-to-fine process, was proposed in [43]. A single template matching can be used to approximate the position of an aircraft object in the coarse step, whereas principal component analysis and kernel density can be applied to estimate the parametric shape in the fine step. A classification algorithm for aircraft recognition with high accuracy was proposed in [36]. Each aircraft image was captured from a great distance from the ground, and then depicted as small objects with multi-textures, orientations and noise. These variations make it extremely difficult to obtain useful features for type recognition process. As a result, image processing was performed such as noise removal, geometric adjustment and rotational correction. In [37], it was suggested that using CNN with a limited number of labelled examples may be a problem since it can lead to overfitting. Thus, the ImageNet challenge was proposed as a method for dealing with a very different type of classification problem. Although their proposed solution was in two stages, they managed to solve the limited-data problem. Their methodology included a novel feature fusion algorithm for dealing with the dimensionality of large data. The soft regularised AdBoost method, proposed in [38], was used for the classification of satellite images. This algorithm links RBF networks as base learners. The synthetic multispectral satellite images and hyperspectral satellite images were considered for the experimental stage. A perfect performance was obtained compared to state-of-art methods that support vector machine in low-to-noise ratio images when the dimension of the input image is relatively high. In [39], a new approach called faster edge region convolutional neural network was proposed to improve the accuracy of building detection and classification. A new method for enhancing the detection of boundaries at detected objects was also suggested. The proposed methodology was trained and evaluated with different datasets, and the results were compared with other methods such as faster region convolution neural network and single-shot multi-box detector. The experimental results exhibited the ability to acquire an average detection accuracy of 97.5% with better resistance to shadow effect. In [40], a new segmentation method was proposed based on the morphological features between connected components in the set of images. The derivation of morphological profile formalises this feature in multiscale segmentation. Multi-segmentation processing is considered when there are complex image scenes, such as satellite images, where a thin envelop or nested regions must be retained. The method used in this paper has the significant benefit of accurately determining the values of the surface area and length of an airplane. We can then determine the type of any detected airplane depending on its actual physical values. This will result in extremely precise flight control at airports. Table 1 provides a summary of related works.

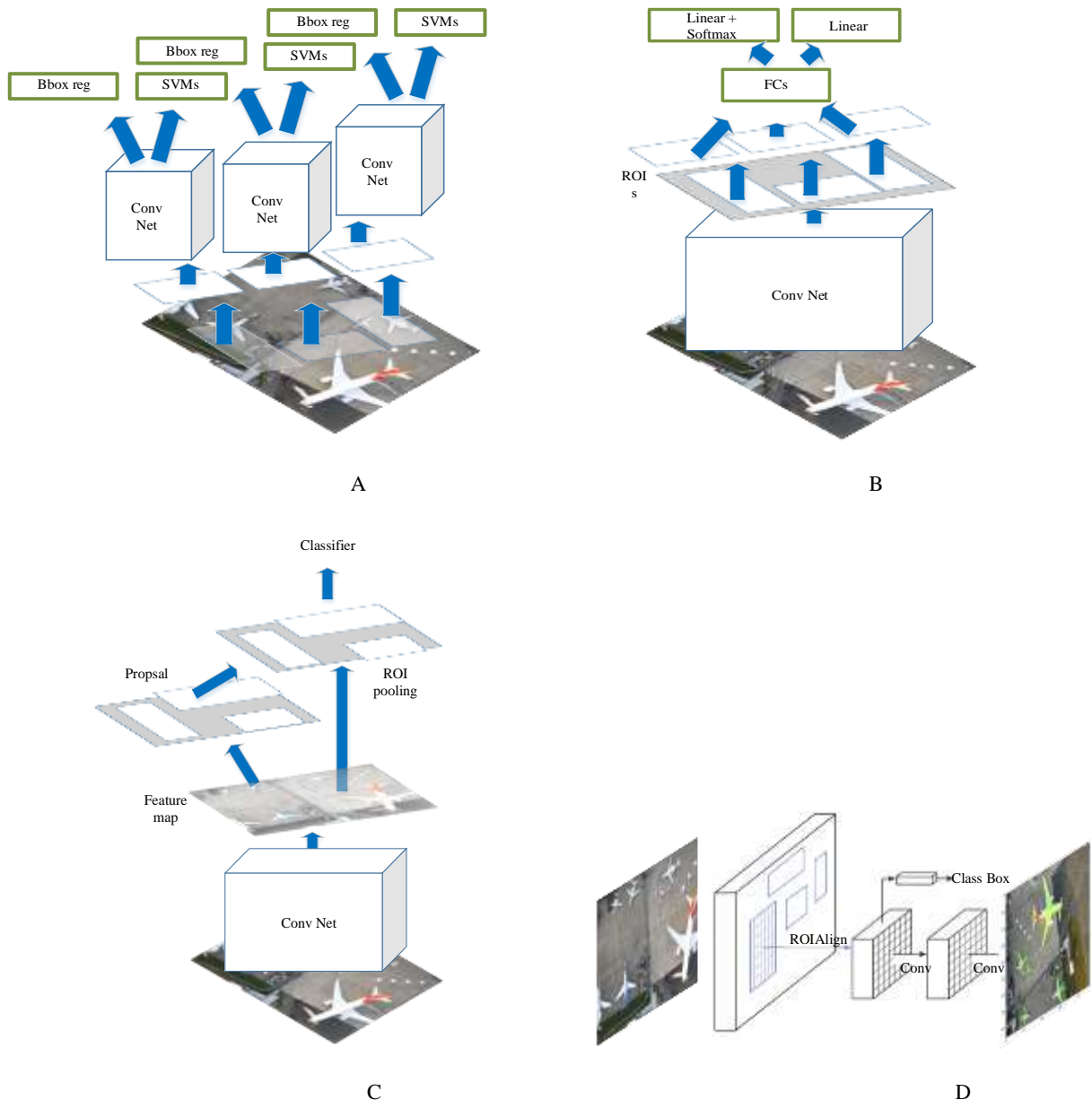

**Figure 4:** Deep learning models a) RCNN b) Fast RCNN c) Faster RCNN d) Mask RCNN [5-8]

### III. SYSTEM AND SIMULATION MODEL

This section focuses on the practical elements of this paper. The first step is the acquisition process to build a dataset using images. All images in the dataset were captured by two eBee classic drones that simultaneously flew over Le Bourget airport in Paris [41]. The drone images are typical images of a simple composition. As a result, they can be directly forwarded to any system without further processing. This is particularly useful for target detection applications. Since there are less optical factors, such as clouds or other atmospheric effects, this type of image has higher resolutions and lower noise than satellite images. Various adjustments (such as picturing moving objects or infrared bodies) can also be made when capturing these images. Other options are available, such as zooming in or using the timer.

**TABLE1 SUMMARY OF RELATED WORK**

| No | Author | Contributions |
|----|--------|---------------|
| 1 | Alganci, Soydas & Sertel | Comparative research on deep learning approaches for airplane detection from very high-resolution satellite images. SSD, faster RCNN and YOLO are employed to detect airplanes using satellite images [33]. |
| 2 | Zhang et al. | The R-CNN-Based Ship Detection from High Resolution applies R-CNN for ships detection using high resolution images [34]. |
| 3 | Lei et al. | Vehicle detection in aerial images is based on region convolutional neural networks and hard negative example mining. They employed HRPN and replaced the classifier by a cascade of boosted classifiers in faster R-CNN [42]. |
| 4 | Yang et al. | Aircraft detection in remote sensing images are based on saliency and convolution neural network. They used a CNN on satellite images to conduct aircraft and vehicle detection. They replaced the selective search by saliency [27]. |
| 5 | Bi et al. | A visual search inspired computational model is used for ship detection in optical satellite images. They accomplished ship detection using SIFT extracted previously to feed SVM using panchromatic images [43]. |
| 6 | Cheng & Han | Learning rotation-invariant convolutional neural networks were used for object detection in VHR optical remote sensing images. They employed recent, SPMK, SSC, SSAE and rotation-invariant CNN to conduct object detections [44, 45]. |
| 7 | Xiang et al. | Vehicle detection in satellite images is accomplished by hybrid deep convolutional neural networks, IEEE Geoscience and Remote Sensing Letters. They used HDNN to carry out vehicle detection [11]. |
| 8 | Xiang et al. | Aircraft detection is conducted by deep convolutional neural networks. They used DBN to detect aircrafts [46]. |
| 9 | Chen, Zhan & Zhang | Geospatial Object Detection in Remote Sensing Imagery is based on Multiscale Single-Shot Detector with Activated Semantics. They used SSD to detect objects [47]. |
| 10 | Yang et al. | Automatic ship detection in remote sensing images from google earth of complex scenes is based on multiscale rotation Dense Feature Pyramid Networks and Remote Sensing. They used R-DFPN-Fast-RCNN-Resnet-101 to accomplish ship detection [48]. |
| 11 | Mundhenk et al. | A large contextual dataset is used for the classification, detection and counting of cars with deep learning. They used YOLO to detect cars [49]. |
| 12 | Zhou et al. | Weakly supervised target detection in remote sensing images is based on transferred deep features and negative bootstrapping. They used AlexNet-CNN to detect airplanes, vehicles and airports [32]. |
| 13 | Zhihuan et al. | Rapid target detection in high resolution remote sensing images is accomplished using the YOLO model of the International Archives of Photogrammetry. They used R-CNN, Fast R-CNN, Faster R-CNN and YOLO to conduct object detection for multiple objects [50]. |
| 14 | Polat & Yildiz | Stationary aircraft detection is taken from satellite images. They used 2D Gabor filter-SVM to handle aircraft detection [51]. |
| 15 | Kang et al. | A modified RCNN based on CFAR is used for SAR to perform ship detection [52]. |
| 16 | Van Etten | Used YOLT to conduct multiple object detection [53-55] |
| 17 | Bashmal et al. | Used Siamese-GAN to perform multiple object detection [56] |
| 18 | Redmon & Farhadi | Used YOLT2 to accomplish multiple object detection [9] |
| 19 | Zalpour et al. | Used Faster R-CNN to handle oil tank detection [29] |
| 20 | Khan et al. | Used Edge Boxes-CNN to conduct aircraft detection [30] |
| 21 | Sun et al. | Used SSCBOW to perform aircraft detection [18] |
| 22 | Körez & Barısçı | Used Faster R-CNN-Resnet50 to manage multiple object detections [57] |
| 23 | Bruzzone & Carlin | A multilevel context-based system is employed for classifying very high spatial resolution images. They proposed a pixel-based system for geometrical spatial images [35]. |
| 24 | Li & Castelli | Texture feature set is derived for content-based retrieval of satellite image database. This work has a benchmark consisting of 37 satellite image chips from different satellite instruments [58]. |
| 25 | Liu et al. | Aircraft recognition in high-resolution satellite images is accomplished using coarse-to-fine shape prior. They proposed a method which integrates the high level information of a shape prior considered as a coarse-to-fine process [59]. |
| 26 | Hsieh et al. | Aircraft type recognition is applied in satellite images. They proposed a classification algorithm for aircraft recognition with high accuracy [36]. |
| 27 | Marmanis et al. | Deep learning earth observation classification is accomplished using ImageNet pretrained networks. They proposed a method that allows pretrained CNN to tackle an entirely different type of classification problem called the ImageNet challenge [37]. |
| 28 | Camps-Valls & Rodrigo-González | Classification of satellite images with regularised AdaBoosting of RBF neural networks is accomplished. They proposed the soft regularised AdBoost method which is used for classifying satellite images [38]. |
| 29 | Reda & Kedzierski | They proposed a new approach called faster edge region convolutional neural network to improve the accuracy of building detection and classification [39]. |

| 30 | Pesaresi & Benediktsson | A new approach was proposed for the morphological segmentation of high-resolution satellite imagery. The derivation of morphological profile formalises this feature in multiscale segmentation [40]. |
| 31 | Emilio Guirado | They explored and analysed the precision of object-based picture examination and Mask RCNN in the segmentation of scattered vegetation in a dryland ecosystem. [60] |
| 32 | Chunming Han | They used Faster RCNN to detect the chimneys in three high-resolution remote sensing images of Google Maps[61] |
| 33 | Ziran Ye | They proposed framework effectively using CNN to mapand discriminate rural settlements[62] |

This type of drone is extremely efficient as it can cover up to 12 km2 with a single mapping flight. Their body structure allows only one option for mounting a camera. The physical structure of these drones enables direct acquisition by looking straight down, pointing to the nadir direction, or directly below the camera. Table 2 provides further details of the camera specifications [27].

Figure 5 introduces a visual representation of the camera sensor and field of view (FOV). Using the camera sensor's width (WS), image width (WI), focal length (FL) and drone altitude (h), the ground sample distance (GSD) or ground resolution can be calculated [33][34]. As illustrated in Figure 5, the field of view is the angle that depends on the focal length and sensor size. The focal length of a lens is its angular field of view; therefore, a shorter focal length signifies a wider angular field. Additionally, the shorter focal length of a lens indicates that a shorter distance is needed to acquire the same focal length compared to a longer focal length. As a result, the focal length of a thin convex lens is measured as the distance between the plane where the image is formed and the back of the lens height. The GSD can be calculated using Equation 1. For a wide range of applications, the required distance from objects and the desired field of view (known as quantities) are computed. These quantities are extremely important since they can be used to determine the required angular field of view.

TABLE 2 CAMERA SPESIFICATIONS.

| Item | Specification |
|---|---|
| Sensor | 1 inch (12.75mm x 8.5 mm) |
| RGB lens | F/2.8-11, 10.6 mm (35 mm equivalent: 29 mm) |
| RGB resolution* | 5,472 x 3,648 px |
| Exposure compensation | ±2.0 (1/3 increments) |
| RGB shutter | Global Shutter 1/30 – 1/2000s |
| White balance | Auto, sunny, cloudy, shady |

$$\text{GSD} = \frac{W_s \times h \times 100}{FL \times W_I} = \frac{12.75\ mm \times 120\ m \times 100}{10.6\ mm \times 4608\ px} = 3.13\ cm/px \quad (1)$$

Taking full advantage of GSD, this paper calculates an approximate surface area (Aappr) for detected planes by multiply the total number of pixels (# px) for each airplane with the GSD value. The total number of pixels for each airplane is determined by the detected masks since masks have tolerances with edge coverage; these calculations are approximated and Aappr is then calculated using Equation 2.

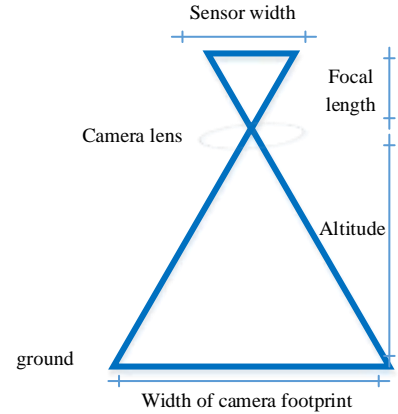

**Figure 5:** GSD calculation parameters

$$\text{GSD} = \frac{W_s \times h \times 100}{FL \times W_I} = \frac{12.75\ mm \times 120\ m \times 100}{10.6\ mm \times 4608\ px} = 3.13\ cm/px \quad (1)$$

$$A_{appr} \cong px \times GSD \quad (2)$$

Due to image resizing, there is a need to rescale the approximate areas to match the original dimensions. For the first seven models, the input images were the size of 304 x 231 px. Thus, the reducing scale is 1:15 for each width and height. The multiplication factor of 152 was consequently applied to each approximate area. Figure 6 illustrates the surface area. detection output. Most airplane manufacturers do not offer precise numbers of their planes' surface area. Instead, they offer the plane's dimensions. Under the airport's database, surface areas for each plane can be calculated. Airports may perform such calculations using data extracted from airplane dimensions or they may have officers with sufficient sense and expertise to compare measured approximated surface area with actual surface area. The advantages of approximated area would be used rather than discarded. Figure 7 illustrates an example of surface area dimensions. Companies normally provide the surface area of wings, although the remaining areas are not calculated. Figure 7 presents several surface area visual detection implementations If the airport has a poor database, then this article offers an additional feature of using length to detect the type of airplane since most manufacturers offer plane length.

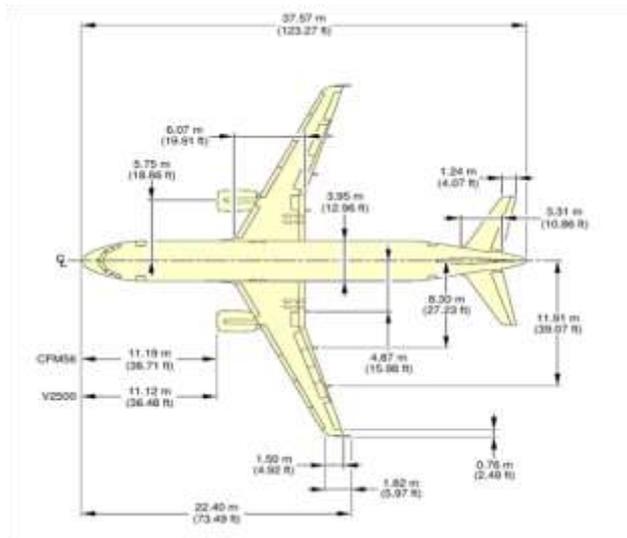

**Figure 6:** Airplane dimensions example-airbus a320 [35].

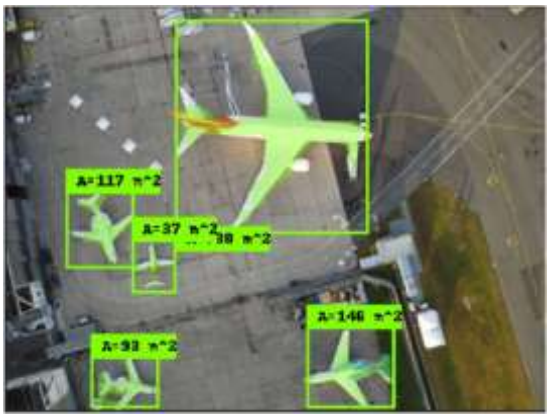

A

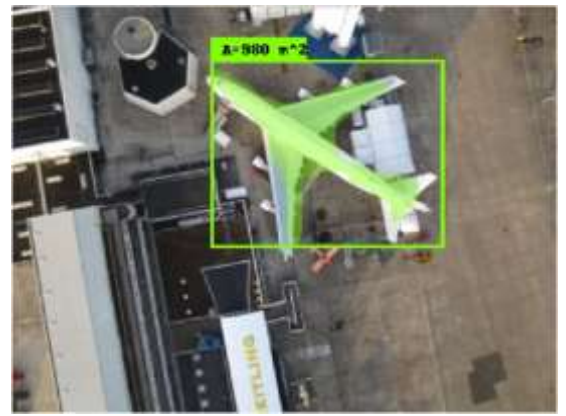

B

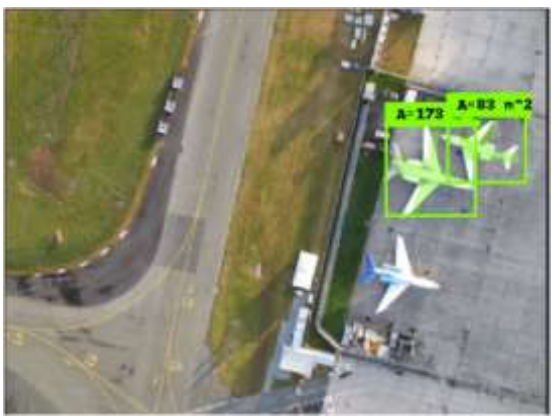

C

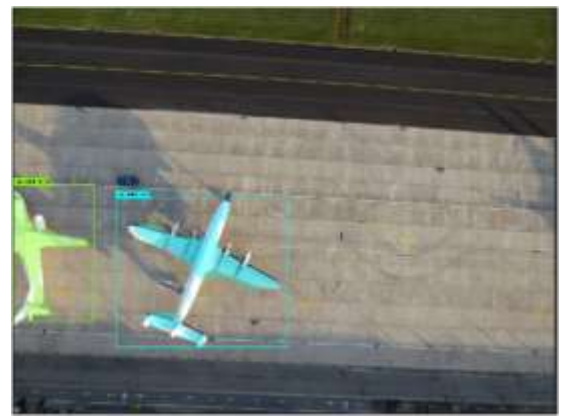

D

**Figure 7:** Surface area detection outputs.

Identifying the length approach is derived from the surface area approach by determining the shortest pairwise distances between pixels within the masked area, then finding the furthest distance. The furthest distance can be computed between pixels located as vertices in a convex hull, thus, using the convex hull algorithm reduces computations for a fewer number of pixels.

The smallest convex polygon that links and surrounds all points within a set of data points in a plane is called the convex hull to this set. The vertices are the points where two consecutive edges meet, and the polygon is an area of a plane bounded by line segments (called edges) that are connected end to end in a cycle. Starting from the leftmost point (the point with the minimum x-coordinate value), wrapping points in a counter-clockwise direction until the same point is reached would be the simplest method.

The ultimate aim of this research is to identify the type of airplane that has been detected. Fortunately, the International Paris Air Show has provided a list of aircraft types to apply for the collected dataset of aircrafts visible in the images. The output lengths can be calculated by calculating the average of the detected length over the original validation set, as seen in Equation 3. The evaluation can then be accomplished by comparing the output lengths to actual aircraft lengths:

$$Detected\ length\ avg = \frac{\sum_{j=1}^{j=n} x_j}{n} \quad (3)$$

where j is the number of validation images, x is the validation image and n is the total number of validation sets used in the evaluation. Table 3 lists the abbreviated names of the airplanes as well as their actual lengths..

TABLE 3 AIRPALNE NAMES, SHORTCUT AND ACTUAL LENGTHS

| Plane Full Name | Plane Shortcut | Actual length |
| --- | --- | --- |
| LockheedMartin-LM100J | LM100J | 35 |
| GULFSTREAM-G-280 | G-280 | 20 |
| GULFSTREAM-G-550 | G-550 | 29 |
| GULFSTREAM-G-650 | G-650 | 30 |
| Cessna-Citation CJ4 | CJ4 | 16 |
| Cessna-Citation M2 | CM2 | 13 |
| Boeing 787-8 | Bo787 | 57 |
| Airbus A-380 | A-380 | 73 |
| Airbus A-320 | A-320 | 38 |

## IV. RESULTS AND DECISIONS

The results of detecting the airplane type are presented in this section. The goal of this paper, as previously mentioned, is to identify the type of each detected airplane based on its surface area and length. The images were taken in the nadir position since these drones only allow the camera to be placed in one position (the downward position). Since the camera is physically mounted inside the drone's body, the nadir acquisition process was ensured for all images. This process depends on the estimated mask for each detected airplane object. Each mask basically contains a number of pixels within the detected object area in the same image. By using simple calculation steps, the surface area can be easily obtained. This feature can be applied to distinguish between airplane types using the value of the surface area. Unfortunately, there are no reference values for the airplanes' surface areas since manufacturers do not clearly provide them. They do provide other details of airplane components which can be used to evaluate the surface area. Rather than performing complex calculations, this paper proposes an alternative solution by analysing the length of each detected airplane object. The idea is to employ the convex hull algorithm to estimate the distance between the two farthest points, which can be used to determine the length of the detected airplane. These estimated values can then be compared with the original manufacturer's reference values, such as the cabin length of an aircraft.

Based on previous research [63], Model 6 achieved the best performance for both training and validation datasets, as illustrated in Table 4. The model's ability to detect airplane objects even from satellite images has been successfully demonstrated using a satellite dataset as a benchmark. Table 4 also presents the COCO metric values. Model 6 was used to execute the detection process and estimate the masks for each detected airplane. The identification process of airplane type can subsequently begin based on these findings.

TABLE 4 The model performance in detection process

| Metric no | Training Dataset | Validation Dataset | Satellite |
| --- | --- | --- | --- |
| 1 | 0.921 | 0.573 | 0.472 |
| 2 | 0.99 | 0.938 | 0.932 |
| 3 | 0.983 | 0.645 | 0.43 |
| 4 | 0.875 | 0.426 | 0.26 |
| 5 | 0.937 | 0.627 | 0.499 |
| 6 | 0.969 | 0.781 | 0.687 |
| 7 | 0.402 | 0.289 | 0.24 |
| 8 | 0.942 | 0.617 | 0.53 |
| 9 | 0.943 | 0.625 | 0.538 |
| 10 | 0.901 | 0.504 | 0.387 |
| 11 | 0.958 | 0.672 | 0.56 |
| 12 | 0.982 | 0.829 | 0.738 |

According to the statistics provided in Table 5, the best length estimation was achieved by Models 2, 4 and 5 with an average accuracy of 89%. Model 6 had approximately the same average, 87%, and performed perfectly in terms of detection. Since Model 3 failed to detect certain airplane types, its results were rejected. To determine the airplane type, the detected length values were matched to the nearest actual lengths. Abbreviated names were used to produce proper output annotations..

TABLE 5 LENGTH DETECTION ACCURACY %

| Plane | Models | | | | | | | |
|---|---|---|---|---|---|---|---|---|
| | 1 | 2 | 3 | 4 | 5 | 6 | 7 | 8 |
| LM100J | 98 | 99 | 98 | 98 | 99 | 99 | 98 | 99 |
| G-280 | 91 | 99 | 93 | 98 | 98 | 99 | 91 | 80 |
| G-550 | 78 | 78 | 73 | 75 | 78 | 78 | 71 | 78 |
| G-650 | 86 | 95 | 92 | 91 | 91 | 90 | 86 | 98 |
| CJ4 | 91 | 97 | - | 96 | 91 | 91 | 85 | 77 |
| CM2 | 77 | 85 | - | 85 | 92 | 73 | 96 | 69 |
| Bo787 | 97 | 99 | 99 | 99 | 95 | 99 | 60 | 96 |
| A-380 | 99 | 98 | 99 | 100 | 99 | 100 | 88 | 95 |
| A-320 | 53 | 53 | - | 56 | 56 | 56 | 43 | 53 |
| Average | 86 | **89** | - | **89** | **89** | 87 | 80 | 83 |

Figure 8 displays the visualisation results of airplane type identification. Figure (8-a) shows the true airplane type classification for Bo787. Figure (8-b) presents the true airplane type classifications for Bo787 and Cm2, but false identification for G-550. Figure (8-c) shows the true airplane type classifications for Bo787, G-550, Cm2, G-280 and CJ4. Figure (8-d) reveals the true airplane type classification for A-380.

In the field of remote sensing research, massive amounts of data should be available. Labelled data is required for image interpretation and target detection, such as airplane detection. Although template matching methods can address this problem, they are difficult to apply at a wide scale. The end-to-end method is a generally reliable method for achieving better behaviour. End-to-end requires vast amounts of labelled data and is considered as the main issue that somehow affects the accuracy. Despite the fact that the number of examples used in the training process is limited, the results are impressive

Overall, the surface areas or lengths can be used to determine the plane type. The surface area of any detected plane is calculated using ground resolution and number of pixels. The actual surface areas of planes which must be calculated are not explicitly provided. Manufactures usually specify the length of an airplane produced, therefore, the key feature for determining the type of any detected plane is to calculate its length using the convex hull algorithm.

The performance of the proposed method has been evaluated by calculating the accuracy of the detected arithmetic mean vs actual values. The obtained accuracy values exhibited promising results. The confusion matrix, also known as the error matrix, was used to assess the performance of the proposed method. Table 6 presents some errors in airplane type identification. These errors are linked to airplanes with similar dimensions, such as the G-550 and G-650, which have a 1 meter difference in length. Another justification is that some images contain cropped airplanes which impair length detections. Therefore, correct location and timing are crucial when capturing images. This can be addressed by scheduling drone flights at precise timing and locations. Additional detection for wing surface area will be added in future works to improve airplane type identification.

TABLE 6 CONFUSION MATRIX

| | LM100J | G-280 | G-550 | G-650 | CJ4 | CM2 | Bo787 | A-380 | A-320 |
|---|---|---|---|---|---|---|---|---|---|
| LM100J | 9 | | | 2 | | | | | 1 |
| G-280 | | 20 | 4 | | 2 | | | | 2 |
| G-550 | | 1 | 10 | 3 | | | | | 1 |
| G-650 | 2 | | 2 | 8 | | | | | 1 |
| CJ4 | | 3 | | | 11 | 3 | | | |
| CM2 | | | | | | 31 | | | |
| Bo787 | | | | | | | 12 | 2 | |
| A-380 | | | | | | | | 6 | |
| A-320 | 2 | | | | | | 1 | 2 | 4 |

## V. CONCLUSION

A reliable technique for airplane type identification based on airplane detection utilizing a mask area convolution neural network was proposed in this study. This approach makes use of surface area and length computations to determine the observed object's size. Because it is generally supplied as one of the manufacturer's specifications data, length is a crucial element in defining the kind of airplane. This study offers a realistic answer to the problem of airport traffic congestion. According to the findings, the suggested method is capable of detecting a wide range of aircraft types. Knowing the type of plane at the airport will provide you important information like physical and technical specs. The physical specification may be used to calculate the amount of space that should be given to each aircraft. The technical specification can be utilized to establish the services that are necessary as well as the time period. Overall, by employing suitable, simple, and low-cost approaches, our study successfully helps to resolving airport traffic congestion issues.

**ACKNOWLEDGMENT**
The authors appreciate the reviewers' valuable time to review this work.

**APPENDIX:**
The abbreviations used are first introduced in the text then summarized as a list below table 7 be more convenient.
[1]

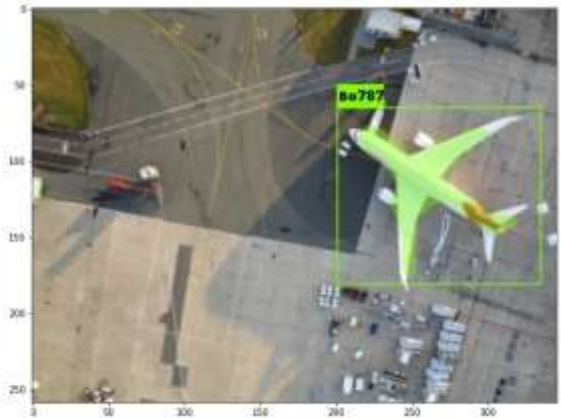 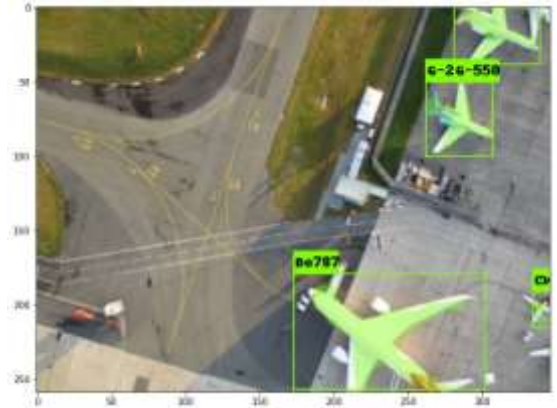

A B

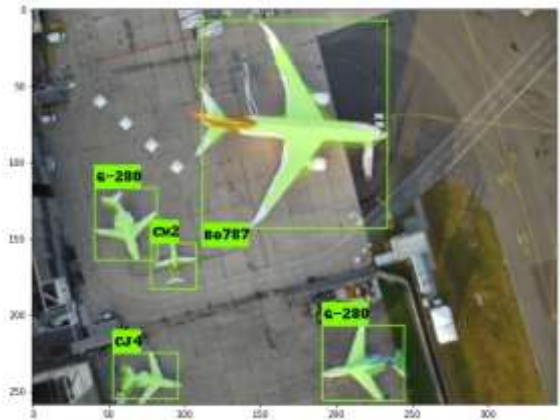 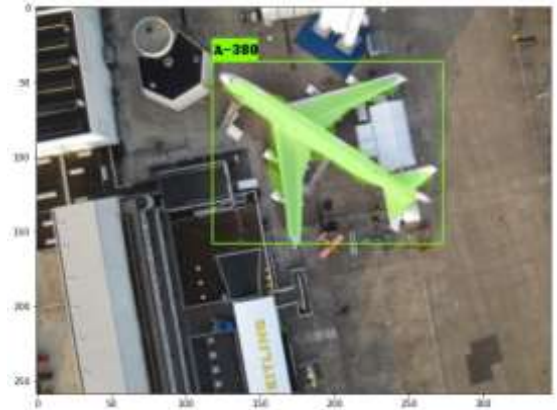

C D

**Figure 8:** Airplane type identification visualization results

TABLE 7  SUMMARIES OF ABBREVIATIONS

| Abbreviation | Meaning |
| --- | --- |
| AP | Average Precision |
| AR | Average Recall |
| CNN | Convolutional Neural Network |
| COCO | Common Objects In Context |
| CPU | Central Processing Unit |
| CUDA | Computer Unified Device Architecture |
| CUDNN | The NvidiaCuda Deep Neural Network Library |
| DBN | Deep Belief Network |
| DL | Deep Learning |
| Faster RCNN | Faster Region-Based Convolutional Network |
| FN | False Negative |
| FP | False Positive |
| GPU | Graphics Processing Unit |
| HDNN | Deep Network |
| HRPN | Hyper Region Proposal Network |
| IoU | Intersection Over Union |
| MAP | Mean Average Precision |
| MAR | Mean Average Recall |
| RAM | Random Access Memory |
| RBM | Restricted Boltzmann Machines |
| RCNN | Region Convolution Neural Network |
| ReLU | Rectified Linear Unit |
| RNN | Recurrent Neural Network |
| RPN | Region Proposal Network |
| RS | Remote Sensing |
| RS | Remote Sensing |
| SGD | Stochastic Gradient Descent |
| SIFT | Scale-Invariant Feature Transform |
| SIFT | Scale-Invariant Feature Transform |
| SVM | Support Vector Machine |
| TN | True Negative |
| TP | True Positive |
| YOLO | You Only Look Once |

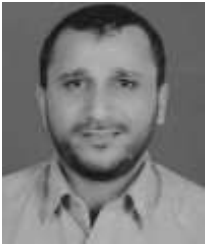

**IBRAHEEM SHAYEA** received the B.Sc. degree in electronic engineering from the University of Diyala, Baqubah, Iraq, in 2004, and the M.Sc. degree in computer and communication engineering and the Ph.D. degree in mobile communication engineering from The National University of Malaysia, Universiti Kebangsaan Malaysia (UKM), Malaysia, in 2010 and 2015, respectively. Since the 1st of January 2011 until 28 February 2014, he has been Research and a Teaching Assistant with Universiti Kebangsaan Malaysia (UKM), Malaysia. Then, from the 1st of January 2016 until 30 June 2018, he joined Wireless Communication Center (WCC), University of Technology Malaysia (UTM), Malaysia, and worked there as a Research Fellow. He is currently a Researcher Fellow with Istanbul Technical University (ITU), Istanbul, Turkey, since the 1st of September 2018 until now. His main research interests include in wireless communication systems, mobility management, radio propagation, and the Internet of Things (IoT).

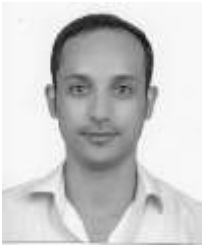

**W.T Alshaibani** received the B.S. degree in electrical engineering from Yarmouk University, Irbid, Jordan, in 2015 and M.Cs in satellite communication and remote sensing from Istanbul technical university (ITU), Istanbul, Turkey. He is currently pursuing the Ph.D. degree in Electrical engineering at ITU. His research interests include 5G and deep learning fields.

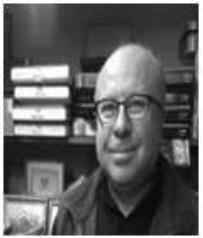

**Mustafa Helvaci** received the B.S., M.S. degree and Ph.D. degrees from Ankara University in 1989, 1994 and 2003 consequently. He works on electromagnetic waves, RF, MW and IR theory, telecommunication, radiation material interactions, radiation and heat transfer, molecular spectroscopy and chemical abundance analysis with remote sensing techniques .

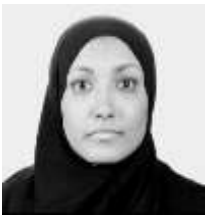

**SAWSAN A. SAAD** received the B.Sc. degree in Electrical Engineering from the University of Khartoum, Sudan, in 2003 and the M.Sc. and the Ph.D. degrees in Mobile Communication Engineering from Universiti Kebangsaan Malaysia (UKM), Malaysia, in 2008 and 2016, respectively. Worked as Asst. Prof. at Future University, Sudan. Currently works as the head of Computer Engineering Department in University of Ha'il, Saudi Arabia. Her research interests include mobile wireless communication systems, heterogeneous networks, mobility management, and Internet of Things (IoT).

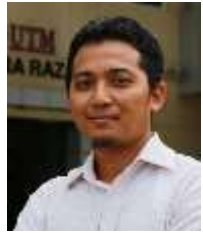

**Azizul Azizan** obtained his B.Eng. (Hons.) Electronics Engineering (majoring in Telecommunications) degree (2002) from Multimedia University, Malaysia. He received his PhD qualification (2009), from University of Surrey, UK in the area of 3.5G physical layer adaptation for satellite systems. He later joined the Malaysian Communication and Multimedia Commission for more than 6 years overseeing spectrum and numbering policies, and telecommunications resource management administration. He is currently with Advance Informatics Department, Razak Faculty of Technology and Informatics, Universiti Teknologi Malaysia where his research areas include Wireless Communications, Edge Computing, Cyber physical Systems, Business Intelligence and STEM Education.

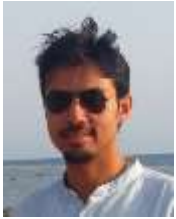

**Fitri Yakub** received his DipEng. and BEng. Degrees in Mechatronics Engineering and Electronics Engineering from Universiti Teknologi Malaysia in 2001 and 2006 respectively. He obtained MSc. in Mechatronics Engineering from International Islamic University Malaysia in 2011. He received doctorate in Automatic Control Laboratory, Tokyo Metropolitan University in 2015. He is now with the Malaysia-Japan International Institute of Technology as a senior lecturer since Oct 2015. He was attached to Alcon Johor (Ciba Vision Sdn Bhd) as industrial intern under CEO Faculty Program from February 2020 until August 2020. He is a senior member of IEEE and member of IET, SAE, and SICE. He was recipient of an Asian Human Resource Fund by Tokyo Metropolitan Government from 2012 until 2015. His field of research interest includes intelligent control, automatic and robust control, and motion control, which related to applications of positioning systems, vehicle dynamics system, and vibration and control systems, IoT, machine learning